\documentclass[conference]{IEEEtran}
\IEEEoverridecommandlockouts

\usepackage[bookmarks=false]{hyperref}
\mathchardef\mathhyphen="2D
\usepackage{todonotes}

\usepackage{cleveref}[2012/02/15]
\crefformat{footnote}{#2\footnotemark[#1]#3}

\usepackage{cite}
\usepackage{amsmath,amssymb,amsfonts}
\usepackage{algorithmic}
\usepackage{graphicx}
\usepackage{textcomp}
\usepackage{xcolor}
\def\BibTeX{{\rm B\kern-.05em{\sc i\kern-.025em b}\kern-.08em
    T\kern-.1667em\lower.7ex\hbox{E}\kern-.125emX}}
    
\usepackage{soul}

\begin{document}

\title{Cross-Domain Car Detection Using Unsupervised Image-to-Image Translation: From Day to Night}

\author{
\IEEEauthorblockN{Vinicius F. Arruda\IEEEauthorrefmark{1}\IEEEauthorrefmark{4}, Thiago M. Paix\~ao\IEEEauthorrefmark{1}\IEEEauthorrefmark{2}, Rodrigo F. Berriel\IEEEauthorrefmark{1}, Alberto F. De Souza, \textit{Senior Member}, \textit{IEEE}\IEEEauthorrefmark{1},\\Claudine Badue\IEEEauthorrefmark{1}, Nicu Sebe\IEEEauthorrefmark{3} and Thiago Oliveira-Santos\IEEEauthorrefmark{1}} 
\IEEEauthorblockA{\IEEEauthorrefmark{1}Universidade Federal do Esp\'irito Santo, Brazil}
\IEEEauthorblockA{\IEEEauthorrefmark{2}Instituto Federal do Esp\'irito Santo, Brazil}
\IEEEauthorblockA{\IEEEauthorrefmark{3}University of Trento, Italy}
\IEEEauthorblockA{\IEEEauthorrefmark{4}Email: viniciusarruda@lcad.inf.ufes.br}
\thanks{This study was financed in part by Coordena\c{c}\~ao de Aperfei\c{c}oamento de Pessoal de N\'ivel Superior – Brasil (CAPES) – Finance Code 001; Conselho Nacional de Desenvolvimento Cient\'ifico e Tecnol\'ogico – Brasil (CNPq) – grants 311120/2016-4 and 311504/2017-5; and Funda\c{c}\~ao de Amparo \`a Pesquisa do Esp\'irito Santo - Brazil (FAPES) – grant 84412844/2018.}}


\maketitle

\begin{abstract}
Deep learning techniques have enabled the emergence of state-of-the-art models to address object detection tasks. However, these techniques are data-driven, delegating the accuracy to the training dataset which must resemble the images in the target task. The acquisition of a dataset involves annotating images, an arduous and expensive process, generally requiring time and manual effort. Thus, a challenging scenario arises when the target domain of application has no annotated dataset available, making tasks in such situation to lean on a training dataset of a different domain.
Sharing this issue, object detection is a vital task for autonomous vehicles where the large amount of driving scenarios yields several domains of application requiring annotated data for the training process.
In this work, a method for training a car detection system with annotated data from a source domain (day images) without requiring the image annotations of the target domain (night images) is presented. 
For that, a model based on Generative Adversarial Networks (GANs) is explored to enable the generation of an artificial dataset with its respective annotations. The artificial dataset (fake dataset) is created translating images from day-time domain to night-time domain. The fake dataset, which comprises annotated images of only the target domain (night images), is then used to train the car detector model. Experimental results showed that the proposed method achieved significant and consistent improvements, including the increasing by more than 10\% of the detection performance when compared to the training with only the available annotated data (i.e., day images).
\end{abstract}


\IEEEpubid{\begin{minipage}[t][10cm][t]{\textwidth}\ \\[10pt] \centering
  {\normalsize\textcopyright 2019 IEEE. Personal use of this material is permitted.  Permission from IEEE must be obtained for all other uses, in any current or future media, including reprinting/republishing this material for advertising or promotional purposes, creating new collective works, for resale or redistribution to servers or lists, or reuse of any copyrighted component of this work in other works.}
\end{minipage}}

\section{Introduction}
\label{Introduction}

Deep learning techniques have enabled the emergence of several state-of-the-art models to address problems in different domains, such as image classification \cite{berriel2017deep,berriel2017automatic}, regression \cite{berriel2017monthly,berriel2018heading}, and object detection \cite{ranik,tfsign}, which is the focus of this work. 
However, these techniques are data-driven, which means that the performance achieved in a test dataset strongly depends on the training dataset. Therefore, the lack of annotated datasets may hinder the training of these models.
Thus, a challenging scenario arises when a high-performing model in one domain (i.e., target domain) is desired, but the model is trained on a distinct, yet analogous, domain (i.e., source domain). In these situations, the target domain and the source domain are very close in semantics, but are very different in appearance. For example, one might be interested in detecting objects (e.g., people, cars, motorcycles) in a specific target domain (e.g., night-time images, rainy images), but only has annotated images from a different domain  (e.g., day-time image, non-rainy images). 
 
The training is difficult not only because of the amount of data that has to be acquired, but also because of the process of annotating them, which requires time and manual effort. To mitigate the problem of annotating the images, several approaches have been proposed in the literature: annotation tools that facilitate the interaction with the user and make this process easier \cite{labelme}; crowdsourcing annotation tools that rely on people to voluntarily annotate the data \cite{su2012crowdsourcing}, which is sometimes a paid service \cite{amazon}; and automatic labeling that makes use of machine learning techniques to extract features \cite{zhang2012review,carneiro2007supervised}. Although there are many techniques to ease this process, the issue remains open and requires further investigation.

In this context, training good object detectors to work across domains is a highly desirable task. Therefore, a method capable of translating images from one domain to the other could help transferring annotations across domains. The emergence of Generative Adversarial Networks (GANs) \cite{goodfellow2014generative} leveraged the building of image generation methods \cite{radford2015dcgan}, which can address the translation problem. This type of network is based on a very popular deep network which works with images, called Convolutional Neural Network (CNN)\cite{cnn}. Recently, image translation methods based on GAN have emerged \cite{isola2017pix2pix} and further advanced performing image translation between distinct domains in an unsupervised manner. For instance, the authors of \cite{zhu2017cyclegan} used the supervised technique proposed in \cite{isola2017pix2pix} to compose a framework, called CycleGAN. Their approach is capable of translating images between two domains in both directions without requiring any paired data (i.e., requiring exactly the same image scenario collected in the two different domains, which might be difficult or impossible in some contexts). With the set of translated images from the source domain and their respective transferred annotations, an object detector could be trained to work in the target domain.

One important application scenario for cross-domain detection arises in the context of self-driving vehicles, where areas occupied by sidewalks, pedestrians, riders, cars, etc., should be properly identified.
However, the endless amount of drivable environments leads to an enormous quantity of domains in which these systems can be employed, such as day, night, snowy or rainy scenarios. Usually, it is easier to find annotated data in one of these domains, e.g., day-time images, but it is essential that these detectors work accurately in all of them, enabling the autonomous system to work all day long regardless of the training conditions. Considering the lack of annotated driving data available within these different driving scenarios, a method for training robust models to detect objects across these highly dynamic conditions is a challenge.

This work takes the problem of car detection on night scenes where annotations are available only for day images as a test case for the proposed technique of improving cross-domain object detection. We addressed this problem because it is an instance of a domain in which it is difficult to obtain annotated datasets. The proposed method requires a set of annotated images in the day-time and a set of night-time images which are assumed not to be annotated. To cope with the lack of annotated training data in the target domain, i.e., night-time, the system benefits from a GAN-based unsupervised image translator in order to assemble an artificial dataset (i.e., fake dataset) whose annotations are directly inherited from the source domain, i.e., day-time images. This allows for improvements on the performance of the car detector in the target domain.

To evaluate the proposed system, several experiments using real-world driving images in day- and night-time domains were conducted. The results show that the model can better detect cars in the night-time domain when it is trained only with the fake-night dataset than when it is trained with the day-time images only. Moreover, training a model on a dataset composed by the day and fake-night datasets resulted in a more effective model than training on each dataset alone, i.e., only on the day images dataset or only on the fake-night images dataset.

The remainder of this paper is organized as follows. The next section presents the related works. Section \ref{Proposed-Method} describes the proposed cross-domain car detection system. The experimental methodology and the obtained results are, respectively, in Sections \ref{Experimental-Methodology} and \ref{Results and Discussions}. Finally,
conclusions and future works are presented in Section \ref{Conclusion}.
\section{Related Work}\label{Related-Work}
Performing vision tasks in unlabeled target domains has been widely studied in the literature \cite{transfersurvey}. Recently, the advent of GAN-based models have boosted works on Unsupervised Domain Adaptation (UDA), which aims to adapt a model trained on a set of images of a common nature, i.e., source domain, to accomplish the same task on images of a different but common nature, i.e., target domain. For example, a coupled generative adversarial network (CoGAN) was proposed in \cite{cogan} for learning a joint distribution of multi-domains at image-level. Addressing the UDA problem, CoGAN was employed in the problem of adapting a digit classifier to a different domain than the training domain. Similarly, the work presented in \cite{unsupervisedpixadaptation} proposed an unsupervised approach that learns a transformation in the pixel space from one domain to another, evaluating it on object and digit classification tasks. While these approaches adapt representations only at image-level, CyCADA \cite{cycada} also considers the feature-level, outperforming the aforementioned approaches.

Although the UDA problem has been extensively investigated, the majority of the works focus on the classification task with few works addressing the problem in the context of object detection. For instance, \cite{chen2018domain} employed unsupervised domain adaptation in the object detection task, tackling the problem of different source-target domains on both image and instance levels. Their approach is based on the Faster R-CNN \cite{fasterrcnn} model where three novel components were introduced: two domain classifiers, (i) one at image-level and (ii) another at instance-level, and (iii) a regularization loss in order to help the network to learn better domain invariant features. Despite the promising results, the evaluation was conducted with source and target domains with very similar appearance, e.g., using as source computer graphics synthesized images that were very close to the target real images. No domain changes considering real-world situations (such as daylight changes) were tested. 

Closely related to our work, the study in \cite{closelygan} proposes an end-to-end training framework integrating a pixel-level domain adaptation based on CycleGAN and an object detection network. This second part is very similar to the Faster R-CNN, adding only an adversarial network used to classify the domain of the input image. This additional network is trained in the same fashion as in \cite{domainadaptationbackpropagation}, leading to the emergence of features that are domain-invariant in the Region Proposal Nerwork (RPN).
Although addressing a similar context of application (i.e., car detection), the evaluation process was again only performed in very similar domains, i.e., the same domains as in \cite{chen2018domain}. The new method presented in \cite{closelygan} improved the results of \cite{chen2018domain} in about 1\%. An additional drawback of this method is the extremely large amount of GPU memory required by the framework for training. Such constraint imposes a need for modern GPUs capable of simultaneously hosting both networks in memory (CycleGAN and Faster R-CNN) during training.

In this work, a method for training a car detector to operate on a night environment is proposed without requiring the annotations of the target domain. In contrast to \cite{closelygan}, our method requires less GPU memory since only one network is trained at a time. In addition, the proposed approach was evaluated in real-world images (from day-time image domain to night-time image domain) and was showed capable of improving results when compared to the training with only the day images (lower-bound baseline), which is relevant when no annotations are available for the target domain.

To the best of our knowledge, the addressed problem was only tackled with deep learning-based methods.
\begin{figure*}[ht]
	\centering
	\includegraphics[width=\linewidth]{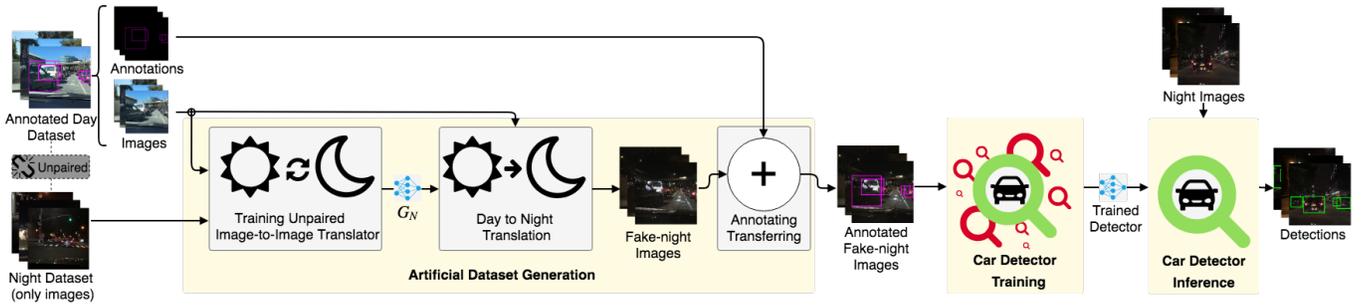}
	\caption{Overview of the proposed system. Firstly, an image-to-image translator model is trained with unpaired day and night images. 
    Then, the day images set is translated to its fake-night versions. The day images annotations is directly transferred to the fake-night images, composing the fake-night dataset. Finally, an object detector is trained resulting in a car detector trained on an image domain that had no annotations previously.}\label{method-overview}
\end{figure*}

\section{Unpaired Image-to-Image Translation for Car detection}
\label{Proposed-Method}

This section describes the method for improving car detection using unpaired image-to-image translation to transfer annotations across domains. The proposed method, illustrated in Figure \ref{method-overview}, comprises three main steps: (i) artificial dataset generation, (ii) car detector training, and (iii) car detector inference. Initially, an unsupervised image-to-image translator is trained using unpaired day and night images with the purpose of generating (fake) night images from day images, i.e., translating the image domain from day to night. The translator is trained to translate only the appearance across domains, which means that the location and pose of the objects of interest (i.e., cars) remain unaltered. Based on this assumption, the annotations (bounding boxes) are directly assigned to the generated images. Therefore, the artificially generated images and their respective annotations comprise together the fake-night dataset. In this second part, an object detector is trained with the generated dataset in order to detect cars in the target domain, where no annotation was previously available. The deployed detector is then ready to infer the cars location in night scenes, i.e., detecting cars in real images from this target domain (inference).

\subsection{Artificial Dataset Generation}

The artificial dataset generation aims to provide a set of annotated images in the target domain that will serve as training data for the detection task in the target domain. The system assumes the availability of annotated day images and non-annotated night images, being two sets of images with $256\times 256$ pixels. The generation process is two-fold. First, a CycleGAN is trained in an unsupervised fashion generating the CNN-based model ($G_N$ in \autoref{cyclegan-overview}) which will be responsible for day-to-night translation. Then, the fake-night images can be automatically labeled using the same annotation as the day images used to train the CycleGAN.

\subsubsection{CycleGAN}

\begin{figure}
	\centering
	\includegraphics[width=0.8\linewidth]{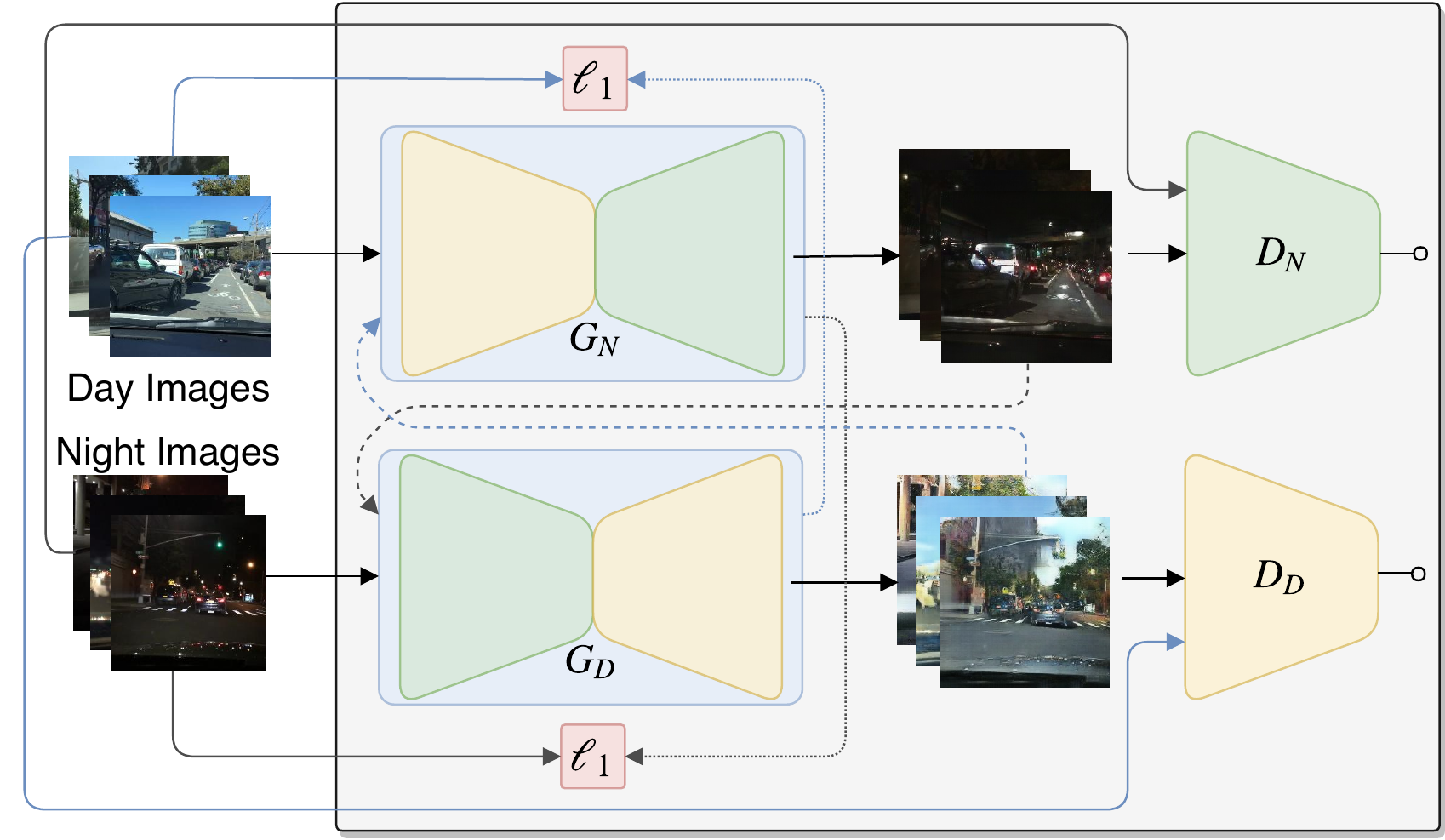}
	\caption{Overview of the CycleGAN framework in this application. $G_N$ maps images from day to night domain, while $G_D$ maps in the opposite way. The discriminators $D_N$ and $D_D$ judge whether an image is a real or fake image in the night and day domain respectively. The cycle-consistency constraint is employed with the $\ell_1$ loss to ensure the reconstruction capability.}\label{cyclegan-overview}
\end{figure}

The CycleGAN framework, illustrated in \autoref{cyclegan-overview}, is trained in a fully unsupervised manner from two \textit{unpaired} (i.e., temporally and spatially detached) set of images, being one in the day domain, and the other in the night domain. The generators $G_N$ and $G_D$ receive the unpaired day and night images respectively, translating them into their own versions in the opposite domain. The $D_N$ discriminator is trained with $G_N$ to correctly distinguish whether a given image is a real night sample or a fake one produced by the $G_N$, which aims to fool $D_N$. Simultaneously, $G_D$ and $D_D$ are trained in the same fashion but with the images in the day domain. To complete the training framework, the fake generated images are fed in the opposite generator in order to try to recover the image in the original domain. This is enforced using a loss that defines a cycle-consistency constraint \cite{zhu2017cyclegan} as $|G_D(G_N(d)) - d|$ and $|G_N(G_D(n)) - n|$, where $d$ and $n$ are real day and night images from the training set, respectively. 

\begin{figure}
	\centering
	\includegraphics[width=0.98\linewidth]{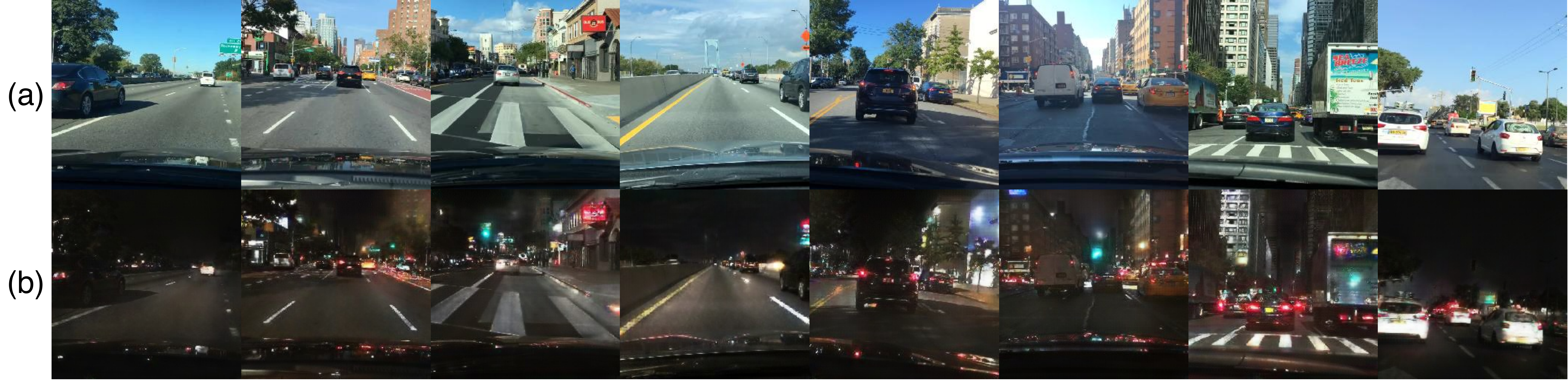}
	\caption{Examples of translated images. The real day training images are shown in (a) and their respective fake-night versions are shown below in (b).}\label{translation-samples}
\end{figure}

Experiments with the cycle-consistency constraint showed that the translation process will mostly not change the global scene structure, as well as the position and geometry of objects (such as cars). Global structure denotes the relationship between the image elements. This fact is exemplified in \autoref{translation-samples}. As it can be seen, the relation between the elements is preserved from the original to the respective fake image.

\subsubsection{Fake Dataset Generation}\label{generating-fake-dataset}

After the training of the CycleGAN, the generator $G_N$ is ready to produce the fake-night images. 
Then, each image of the initial training dataset belonging to the day domain (i.e., day images dataset) is fed into $G_N$, generating a new and corresponding fake-night image.

\begin{figure}
	\centering
	\includegraphics[width=0.98\linewidth]{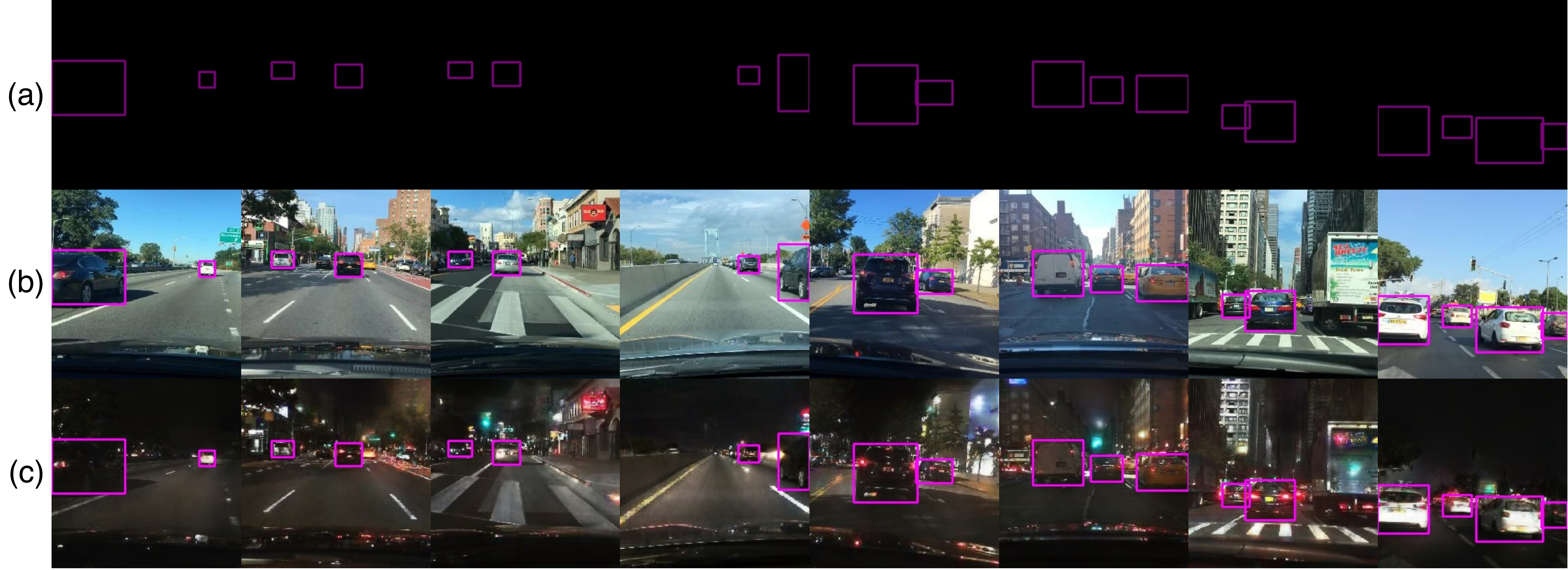}
	\caption{The annotation transfer process. (a) The bounding box annotations of (b) the real day samples are transferred to (c) their respective fake-night versions.}\label{translation-samples2}
\end{figure}

Assuming structural consistency between the source and target images, as was empirically observed (\autoref{translation-samples}), the annotation of the source image (day) can be directly replicated to the target image (night). \autoref{translation-samples2} shows the transfer of bounding box annotation from the day images to their respective night images, i.e., those images generated by the $G_N$ model. The collection of the generated fake images and respective annotations comprise the \emph{fake-night} dataset.

It is important to notice that the CycleGAN does not have to generalize the translation for other images that are not in the training dataset, because it is only used to generate the respective fake-night dataset that is paired with the real day dataset. Once the fake-night dataset is ready, i.e., generated by the translation process, the CycleGAN model is no longer necessary and can be discarded.

\subsection{Car Detector Training}

The car detector uses a general purpose object detector to find cars in the images. An object detector usually receives one image as input and outputs a set of bounding boxes (coordinates of two points in an image defining a rectangle) representing each of the detected objects. Object detectors are usually trained with samples of images annotated with the object of interest. Since this work is interested in studying the detection of cars in night scenes without specific annotation for the night domain, the fake-night dataset produced in the previous step is used as training data. In this work, the Faster R-CNN \cite{fasterrcnn} was adopted as the framework for object detection. Although many other models exist \cite{yolo, retinanet,fastrcnn, rcnn}, this model was chosen as a proof of concept considering its consolidation in the literature, effectiveness and satisfactory performance.

The Faster R-CNN, as originally proposed \cite{fasterrcnn}, comprises two networks that share the same feature maps, being a network responsible for proposing regions of potential objects, and the other for refining and classifying each of the proposed regions. The training of the Faster R-CNN for this problem requires a set of annotated images with bounding boxes of the cars. 

\subsection{Car Detector Inference}

Once the Faster R-CNN model is trained, it can be finally used to detect cars in the target domain. Given a real night image, the trained model predicts bounding boxes of the cars, as well as the confidence level (real-value ranging from 0 to 1) of each detection. Since the CycleGAN is only used to generate data for training the object detector, the computational performance of the inference (i.e., inference time per image) depends only on the chosen object detector (in the case of this study, Faster R-CNN).
\section{Experimental Methodology}
\label{Experimental-Methodology}

This section describes the methodology and materials used in the experiments. 
First, the datasets used to train and evaluate the system are presented. Second, the metric used for quantitative evaluation is described, followed by the discussion of the performed experiments. Subsequently, the training descriptions of the models employed are detailed. Finally, the machine setup used for the experimentation is presented.

\subsection{Datasets}

The Berkeley Deep Drive (BDD) dataset \cite{bdd} was used to train and evaluate the proposed system. This dataset is composed of images ($1280\times 720$ pixels) coming from driving videos across different periods of the day, weather conditions, and driving scenarios. The images of this dataset come with several types of annotations, such as bus, traffic light, traffic sign, person, bike, truck, motor, car, train, and rider, and also drivable area as well as lane marking for driving guidance. The BDD dataset also provides some attributes for each image, such as \textit{time of day}: daytime, night and dawn/dusk; \textit{weather}: rainy, snowy, clear, overcast, partly cloudy and foggy; and \textit{scene}: tunnel, residential, parking lot, city street, gas stations and highway.

Since this work focuses on day-to-night translation, the BDD dataset was filtered based on the \textit{time of day} attribute, keeping only day and night images. Some annotations in the dataset were wrong, e.g., day images annotated as night images and vice-versa, requiring a visual inspection. A further refinement was applied choosing only images whose \textit{weather}'s attribute was `clear' or `partly cloudy' and \textit{scene} being `highway', `city street' or `residential'. These refinements helped to obtain two distinct and homogeneous domains and to reduce possible variability due to the interference from another domain in the dataset. From the object detection annotations, only the car annotations were used. The images were filtered to ensure they all had at least one car.  

To cope with the high processing time imposed by GANs, the images were reduced to $256\times 256$ pixels following two steps: (i) cropping a square of $720\times 720$ pixels positioned in a way that the car's lane was centered, and (ii) rescaling the cropped image to $256\times 256$ pixels. However, the reduction of size made small cars even smaller, which hindered their visual identification. To avoid these situations, cars with the bounding boxes having one of the sides smaller than $20$ pixels in the resized image were removed from the annotations. The occluded or truncated cars (these annotations are also available in the dataset) were removed considering bounding boxes having one of the sides smaller than $30$ pixels.

In total, 12000 images were randomly sampled from the remaining collection, being equally divided (3000 for each) into four subsets: (i) $day_{train}$, used as real images of the source domain for training, (ii) $day_{test}$, used as ground truth of the source domain, (iii) $night_{train}$, used as real images of the target domain for training, and (iv) $night_{test}$, used as ground truth of the target domain. 

To allow the replication of the experiments the Python code to generate the dataset was made publicly available\footnote{\url{https://github.com/LCAD-UFES/publications-arruda-ijcnn-2019/blob/master/README.md}}.

\subsection{Experiments}
To evaluate the proposed method, a set of experiments were performed. The CycleGAN was first used to generate the fake images and the Faster R-CNN was later trained to detect cars. However, the training of methods based on GANs may be very unstable and may leave the optimization process stuck or even diverge \cite{ganconvergence,ganstability,improvedgan}. Due to this inconvenience, the CycleGAN training was repeated a few times until a model capable of producing images with visual appearance closer to real ones was achieved. Once obtained, the fake-night dataset was generated and used for all of the experiments described below.

Different types of training were performed with the Faster R-CNN considering five different datasets: $day_{train}$, $fake\mathhyphen night_{train}$, $day_{train}\cup fake\mathhyphen night_{train}$ (the order can be exchanged depending on the emphasis of the experiment, e.g., $fake\mathhyphen night_{train}\cup day_{train}$), $night_{train}$ and $day_{train}\cup night_{train}$. Each type of training was repeated 10 times for a posterior statistical analysis resulting in a total of 50 models. The difference between the runs on a same training type is the seed for random-based processes, such as weight initialization of the networks and the order in which the images of the dataset are presented to the training.

To evaluate the effectiveness of the proposed method, the analysis was divided in two scenarios of experiment: one considering an object detector that will work throughout the day (i.e., mixing source and target domains) $day_{test}\cup night_{test}$, and one considering an object detector that will work only in the night (i.e., only in the target domain) $night_{test}$.

The experiment evaluating the models on the $day_{test}\cup night_{test}$ resembles the more challenging real-world application problem, in which the system is required to work during the whole day. In this experiment, the lower- and upper-bound baselines are the models trained on $day_{train}$ and $day_{train}\cup night_{train}$, respectively. It is important to note that the baselines training are performed using the full dataset annotation, which includes both images and bounding box annotation. It is assumed that models trained on images from both domains should outperform models trained on day images solely. One hypothesis of this work is that the information of the fake-night dataset can help the detection model to perform better than the lower-bound approaching the upper-bound. To prove the hypothesis, models trained with $day_{train}\cup fake\mathhyphen night_{train}$ were compared to the lower- and upper-bounds.

The experiment evaluating the models on the $night_{test}$ addresses the less challenging real-world application problem, in which the system is required to work during the night. In this experiment, the lower- and upper-bound baselines are the models trained on $day_{train}$ and $night_{train}$, respectively. Again, it is important to note that the baselines training are performed using the full dataset annotation. It is assumed that models trained on target domain should outperform models trained on images of the source domain solely. Another hypothesis of this work is that the information of the fake-night dataset can improve the performance of the model on the target domain. To prove the hypothesis, models trained with $fake\mathhyphen night_{train}$ and $fake\mathhyphen night_{train}\cup day_{train}$ were compared to the lower- and upper-bounds.

\subsection{Performance Metric}

The final purpose of the proposed system is to detect cars accurately. To quantify the quality of the detector, the mean Average Precision (mAP) was adopted following the definition proposed in the PASCAL VOC 2012 challenge \cite{pascal-voc-2012}.

The Average Precision (AP) is defined as the area under the precision-recall curve of a certain object class. Firstly, the curve is built by calculating the precision and recall values of the accumulated true positives or false positive detections. For this, detections are ordered by their confidence scores, and precision and recall are calculated for each accumulated detection. Secondly, interpolated precision values are measured for all recall levels. For this, for each recall level $r$, it is taken the maximum precision whose recall value is greater or equal than $r + 1$. Thirdly, AP is calculated as the total area under the interpolated precision-recall curve.  Finally, the mAP is calculated as the mean of the AP of all classes (in this work there is only the car class).

\subsection{Training Setup}

\subsubsection{CycleGAN}
The architecture used was the same as in the original paper, except for the copy and crop mechanism \cite{unet} that was disabled. The adopted source code is publicly available\footnote{\url{https://github.com/vanhuyz/CycleGAN-TensorFlow}} and was recommended by the authors as an alternative to the original implementation. The CycleGAN was trained with 100 epochs (empirically defined) with one image per batch. The default values were used on the other hyper-parameters.

\subsubsection{Faster R-CNN}
A public source code\footnote{\url{https://github.com/endernewton/tf-faster-rcnn}} was used for carrying out the experiments. The Faster R-CNN feature extractor was initialized with the ResNet-101 \cite{residualnet} weights, which was trained on the ImageNet dataset \cite{imagenet}. This pre-trained model was downloaded from the TensorFlow website\footnote{\url{http://download.tensorflow.org/models/resnet_v1_101_2016_08_28.tar.gz}}. Anchor scales and ratios were defined considering the application working range as $[4,8,16,32]$ and $[0.5,1,2]$, respectively. 

The remaining parameters were defined empirically. The same learning rate was kept for the first 70k iterations and linearly decaying the rate to zero over the next 30k iterations, resulting in 100k iterations with one image per batch. During the training, data-augmentation was performed by flipping the images horizontally.

\subsection{Experimental Platform}
The experiments were carried out in an Intel Xeon E5606 2.13 GHz $\times$ 8 with 32 GB of RAM, and 1 Titan Xp GPU with
12 GB of memory. The machine was running Linux Ubuntu
16.04 with NVIDIA CUDA 9.0 and cuDNN 7.0 \cite{chetlur2014cudnn} installed.
The training and inference steps were done using the TensorFlow framework \cite{tensorflowcite}. The training sessions took, on average, 25 hours for the CycleGAN model and 7.5 hours for the Faster R-CNN model.
In the used setup, CycleGAN translates images at an approximate rate of 6 frames-per-second (fps), whereas Faster R-CNN performs detections at more than 7 fps.

\section{Results and Discussions}
\label{Results and Discussions}

\begin{figure}
	\centering
	\includegraphics[width=\linewidth]{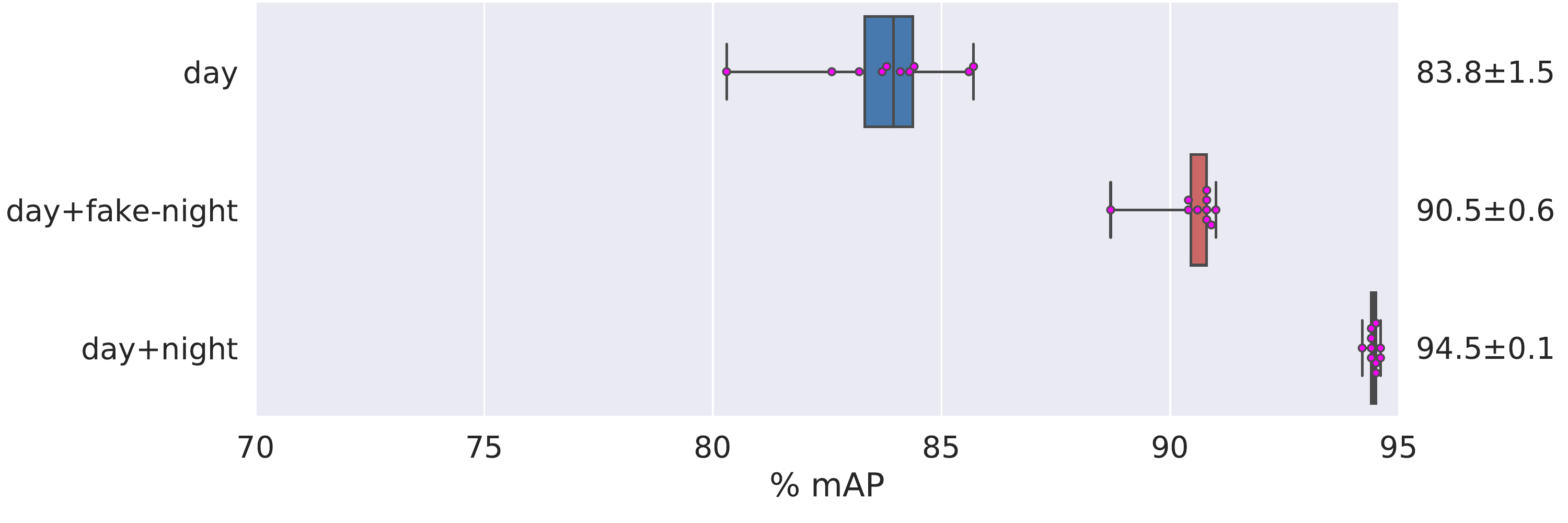}
	\caption{Results of the experiments conducted on $day_{test}\cup night_{test}$. Each dataset used for training is shown in the left vertical axis, whereas the average and standard deviation of the mAP of the 10 runs are in the right vertical axis. The horizontal axis show the actual mAP value. }
	\label{testdaynight}
\end{figure}

The results of the experiment evaluating the proposed method in a more challenging real-world application, i.e., testing on the $day_{test}\cup night_{test}$ dataset, are presented in \autoref{testdaynight}. The results confirm that training the detector in both domains yields better models then training on day images only, with a difference of 10.7\% in the average of the mAP of 10 runs (from now on, in average mAP). Furthermore, the results show that our hypothesis was correct, i.e., that the information of the fake-night dataset aggregated to the training process improves the performance when compared to the lower-bound (training with day images solely). The results show an improvement of almost 7\% in average mAP. In addition, the standard deviation decreased about 60\% indicating that a more robust model was achieved. Moreover, adding the fake-night dataset to the training process yields a model closer to the upper-bound than to the lower-bound, with a difference to the upper-bound of 4\% in average mAP. 

\begin{figure}
	\centering
	\includegraphics[width=\linewidth]{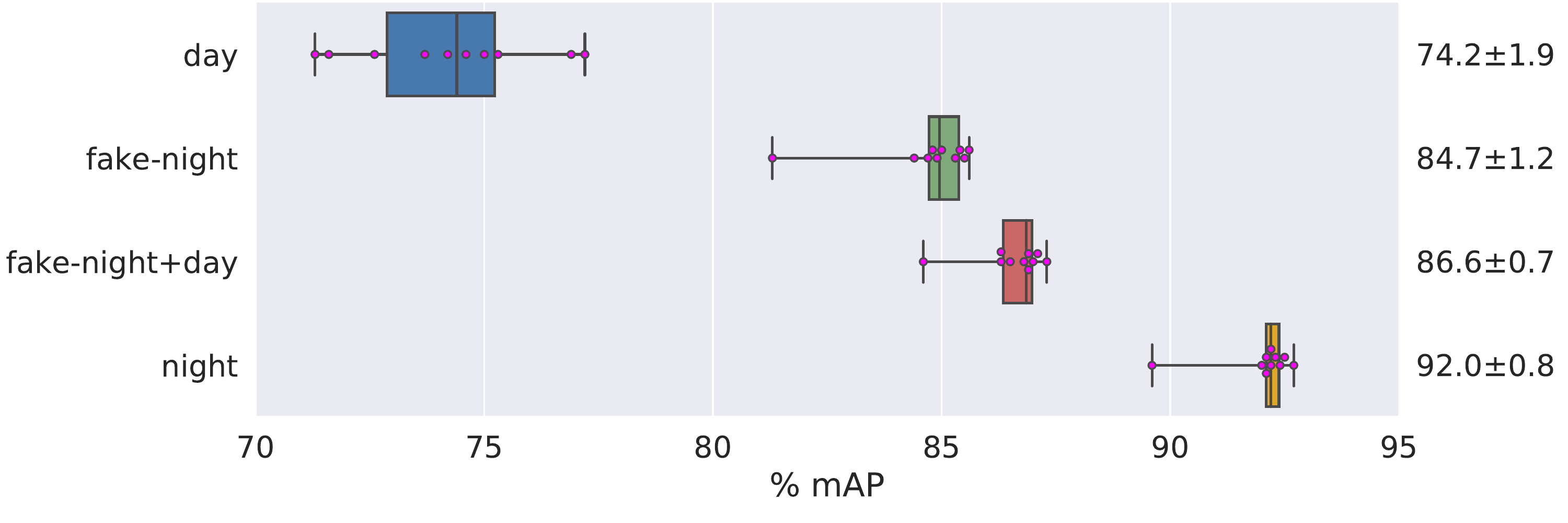}
	\caption{Results of the experiments conducted on $night_{test}$. Each dataset used for training is shown in the left vertical axis, whereas the average and standard deviation of the mAP of the 10 runs are in the right vertical axis. The horizontal axis show the actual mAP value.}
	\label{testnight}
\end{figure}

The results of the experiment evaluating the proposed method in a less challenging real-world application, i.e., testing on the $night_{test}$ dataset, is presented in \autoref{testnight}. In this scenario, two methods were evaluated, with the $fake\mathhyphen night_{train}$ and with $fake\mathhyphen night_{train}\cup day_{train}$. Once more, the results with $fake\mathhyphen night_{train}$ confirmed that models trained with data in the target domain achieve better performance than training on data of the source domain (with a difference of 17.8\% in average mAP). Furthermore, the results showed that our hypothesis was correct, i.e., that the information of the fake-night dataset is more relevant than the day images dataset only (lower-bound). The detector trained with only fake-night images achieved 84.7\% in average mAP, which is 10.5\% greater than the result obtained when training in the day dataset only. Moreover, the results indicate that training with the fake-night dataset only results in a model that is closer to the upper-bound than to the lower-bound, with a difference to the upper-bound of 7.3\% in average mAP. 

The results with $fake\mathhyphen night_{train}\cup day_{train}$ show that fake-night dataset can be used to augment the lower-bound dataset for training. It results in an improvement of 12.4\% in average mAP when compared to the lower-bound. In addition, the results show an improvement of 1.9\% in average mAP when compared to the training with the fake-night dataset only, i.e., augmenting the training data with the day images seems to improve the results. These results indicate that the generation and use of the fake-night dataset brings complementary information to the real day images dataset which results in a better model.

Corroborating with the presented results, employing the Student's t-test (unpaired and two-tailored) pairwise with both lower- and upper-bound baselines for each experiment, the certainty about the acquired results was affirmed with at least 99.9\% confidence.

\begin{figure}
	\centering
	\includegraphics[width=\linewidth]{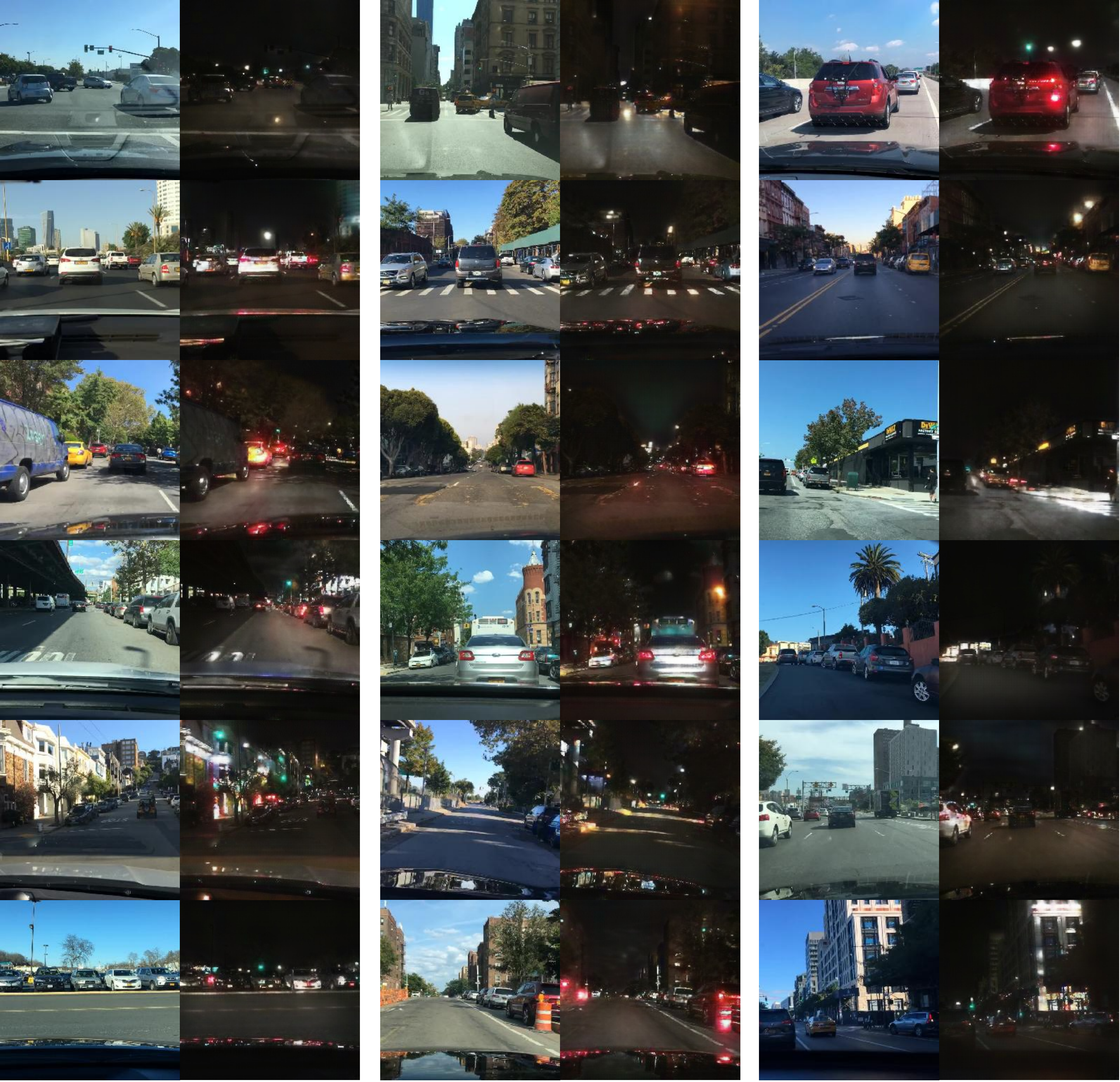}
	\caption{Examples of the $day_{train}$ dataset with their corresponding translations to compose the $fake\mathhyphen night_{train}$ dataset. Many of the generated night images present artifacts that may be seen as unrealistic or fanciful, for example, the image in the third row on the first column illuminated the tree with light dots. Another example is the top image in the third column, where the clouds also became illuminated.}
	\label{fig:sample-unreal-fake-night}
\end{figure}

Qualitative results of the translations are presented in \autoref{fig:sample-unreal-fake-night}. The figure depicts some day-to-night translations, i.e., real day images and their fake-night counterparts. As can be seen, some artifacts are present in the fake images, but the overall appearance of the images looks good. Although the artifacts can be disturbing for models that try to achieve very realistic images, the quantitative results show that the fake images do not have to be perfect to improve the performance of the detections models. However, one could conjecture that better fake images could generate better results. 

\begin{figure}
	\centering
	\includegraphics[width=\linewidth]{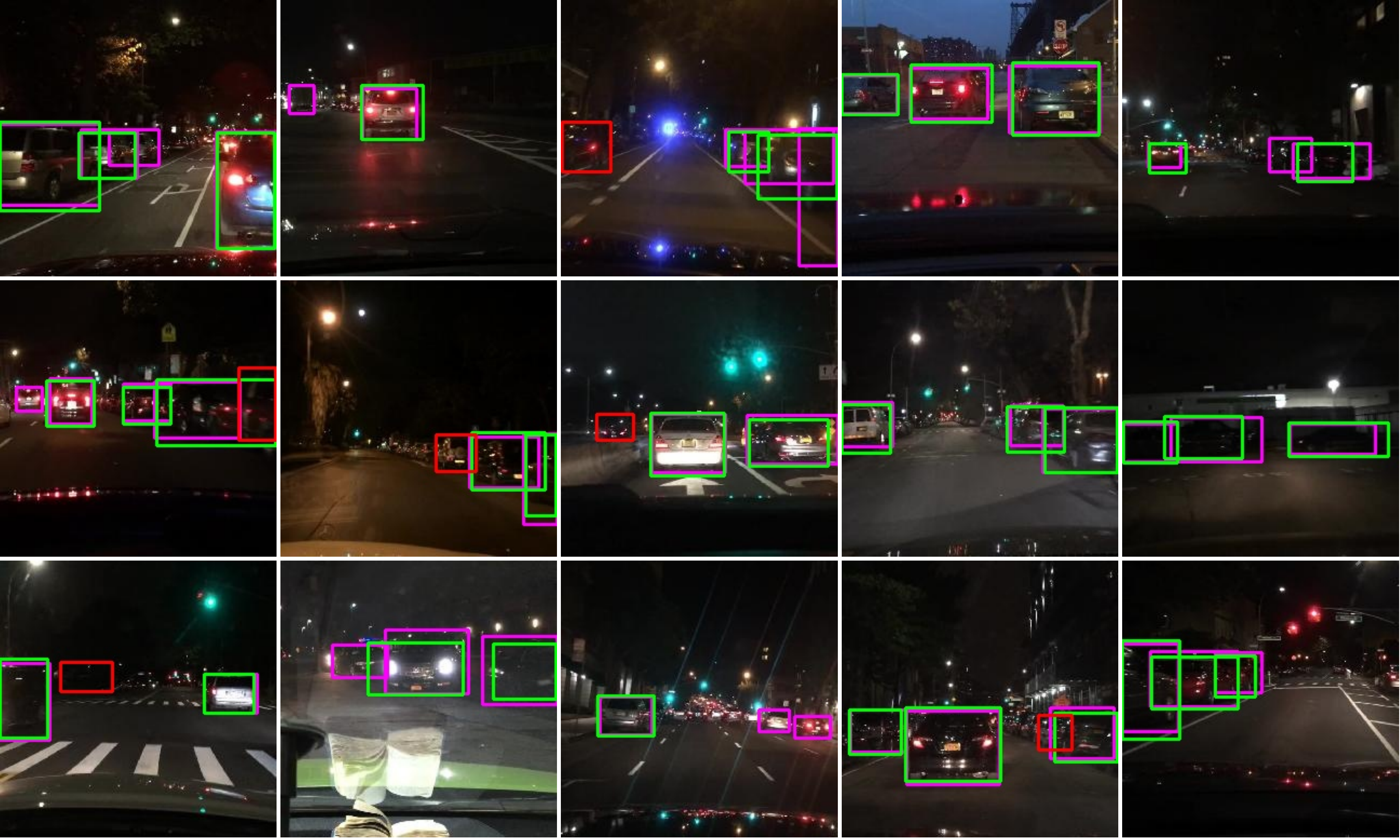}
	\caption{Examples of detections performed by the proposed method trained on the $fake\mathhyphen night_{train}$ dataset. The ground-truth and detections are shown in magenta and green, respectively. The red bounding boxes depict the false-positive detections. Note that, in some cases, the detection contains a car, however, the absence of ground-truth annotation causes it to become a false-positive (e.g., in the second row in the second column).}
	\label{fig:bbox_test}
\end{figure}

\autoref{fig:bbox_test} shows some detections on real night images resulting from training on the $fake\mathhyphen night_{train}$ dataset. As can be seen, most of the detections are as expected, i.e., close to the ground-truth. Some wrong detections (false-positives) can also be seen, nevertheless, some of them are just due to missing ground-truth annotations. A video made publicly available\footnote{\url{https://github.com/LCAD-UFES/publications-arruda-ijcnn-2019/blob/master/README.md}} shows all the detections performed on the $night_{test}$ and $day_{test}\cup night_{test}$ datasets.

\section{Conclusion}
\label{Conclusion}
This work investigated cross-domain (day-to-night) car detection using training datasets without annotations in the target domain (night). To address this problem, we proposed a method to generate a dataset of artificial images annotated automatically to train an object detector in the desired domain. 

To evaluate our proposed method, an investigation was carried out with two experiments considering real-world scenarios. The first experiment investigated the performance of the proposed method when considering detector aiming at working in both domains (day and night). Results showed that augmenting the annotated training data of the source domain (i.e., day images) with annotated artificially-generated images from the target domain (i.e., fake-night) improves the performance, achieving 90.5\% $\pm$ 0.6 in comparison to 83.8\% $\pm$ 1.5 in average mAP. The improvement brings the performance closer to the upper-bound, i.e., to the model trained with real annotated data of both domains. The second experiment investigated the performance of the proposed method when considering detector aiming at working only in a target domain (night) that is different from the source domain with annotated data available (day). Results showed that training the model with annotated artificially-generated images from the target domain (i.e., fake-night) improves the performance when compared to the model trained with the available data of the source domain only, achieving 84.7\% $\pm$ 1.2 in comparison to 74.2\% $\pm$ 1.9 in average mAP. In addition, the results of this second experiment showed that augmenting the artificially-generated images with the source domain data improves the performance in the target domain, improving on 1.9\% in average mAP. One can conjecture that the datasets hold complementary information about the problem.

Both experiments indicated that the proposed method outperformed their respective lower-bounds, showing their success on improving cross-domain object detection using unsupervised image-to-image translation. In addition, the proposed method has the advantage of not having to be able to generalize the translations, i.e., being capable of translating day images to night images outside the training dataset. With this in mind, the translation method does not have to generate good quality images when applied to images outside the training set. Moreover, the results demonstrated that the method can profit from the cross-domain translation even when the translated images are not perfect and show some unwanted artifacts.

Future work should investigate the performance of the method with other GAN-derived models and verify the effect of improving the quality of the fake images in the final detection result. As image translation has become a trending area of research, several methods are emerging with better qualitative results, making them candidates to be employed in the proposed method in future work. Likewise, other state-of-the-art object detectors must be tested, such as YOLO \cite{yolo} and RetinaNet \cite{retinanet}, but are beyond the proof of concept of this work. Finally, to ensure the robustness and the ability to generalize, the method presented here should be evaluated in other scenarios with several distinct domains of detection tasks. 

\section*{Acknowledgment}
We gratefully acknowledge the support of NVIDIA Corporation
with the donation of the Titan Xp GPU used in this
research.


\end{document}